\pdfoutput=1

\documentclass[11pt]{article}

\usepackage[]{ACL2023}

\newcommand{\cmark}{\checkmark}%
\newcommand{\xmark}{ }%

\newcommand{\ours}{PU-ADKA}

\definecolor{c1}{cmyk}{0,0.6175,0.8848,0.1490}
\definecolor{c2}{cmyk}{0.1127,0.6690,0,0.4431}
\definecolor{c3}{cmyk}{0.3081,0,0.7209,0.3255}
\definecolor{c4}{cmyk}{0.6765,0.2017,0,0.0667}
\definecolor{c5}{cmyk}{0,0.8765,0.7099,0.3647}

\definecolor{gg}{HTML}{0F9D58}
\definecolor{rr}{HTML}{DB4437}
\definecolor{bb}{HTML}{4285F4}

\newcommand{\std}[1]{\text{\scriptsize{(#1)}}}

\usepackage{hyperref}
\usepackage{nameref}

\usepackage{times}
\usepackage{latexsym}
\usepackage{graphicx}
\usepackage{algorithm, algorithmic}
\usepackage{amsmath}
\usepackage{booktabs}
\usepackage{tabularx}
\usepackage{colortbl}
\usepackage{hhline}
\usepackage{amssymb} 
\usepackage{array} 
\usepackage{multirow} 
\usepackage{enumitem}

\usepackage{graphicx}
\usepackage{booktabs}
\usepackage{amssymb} 
\usepackage{xcolor}  
\usepackage[T1]{fontenc}
\usepackage[most]{tcolorbox}
\newtcolorbox{prompt}[1]{
    enhanced,
    drop shadow=black!5!white,
    left=4mm,
    right=4mm,
    top=2mm,
    bottom=2mm,
    boxsep=0mm,
    rounded corners,
    title=#1,
    fontupper=\footnotesize\linespread{0.9}\fontfamily{lmr}\selectfont,
    }

\usepackage[utf8]{inputenc}

\usepackage{microtype}

\usepackage{inconsolata}

\title{Active Domain Knowledge Acquisition with 100-Dollar Budget: Enhancing LLMs via Cost-Efficient, Expert-Involved Interaction in Sensitive Domains}

\author{
\textbf{Yang Wu}\textsuperscript{$\clubsuit$} \quad
\textbf{Raha Moraffah}\textsuperscript{$\clubsuit$} \quad
\textbf{Rujing Yao}\textsuperscript{$\heartsuit$} \\
\textbf{Jinhong Yu}\textsuperscript{$\clubsuit$} \quad
\textbf{Zhimin Tao}\textsuperscript{$\diamondsuit$} \quad
\textbf{Xiaozhong Liu}\textsuperscript{$\clubsuit$}\thanks{\, Corresponding author.} \\
\textsuperscript{$\clubsuit$}Worcester Polytechnic Institute \quad
\textsuperscript{$\heartsuit$}Nankai University \quad
\textsuperscript{$\diamondsuit$}Jiangsu University \\
\texttt{\{ywu19, rmoraffah, jyu7, xliu14\}@wpi.edu} \\
\texttt{rjyao@mail.nankai.edu.cn}\quad
\texttt{jsutao@ujs.edu.cn}
}

\begin{document}
\maketitle

\begin{abstract}

Large Language Models (LLMs) have demonstrated an impressive level of general knowledge. However, they often struggle in highly specialized and cost-sensitive domains such as drug discovery and rare disease research due to the lack of expert knowledge. In this paper, we propose a novel framework (PU-ADKA) designed to efficiently enhance domain-specific LLMs by actively engaging domain experts within a fixed budget. Unlike traditional fine-tuning approaches, PU-ADKA selectively identifies and queries the most appropriate expert from a team, taking into account each expert's availability, knowledge boundaries, and consultation costs. We train PU-ADKA using simulations on PubMed data and validate it through both controlled expert interactions and real-world deployment with a drug development team, demonstrating its effectiveness in enhancing LLM performance in specialized domains under strict budget constraints. In addition to outlining our methodological innovations and experimental results, we introduce a new benchmark dataset, CKAD, for cost-effective LLM domain knowledge acquisition to foster further research in this challenging area.\footnote{https://github.com/YANGWU001/PU-ADKA.}

\end{abstract}

\section{Introduction}
Recent advancements in large language models (LLMs) have led to impressive performance gains across a wide range of tasks \citep{naveed2023comprehensive, pal2024domain,yao2025elevating}. However, these gains are not uniformly observed across all domains. In highly specialized and cost-sensitive fields, such as drug discovery and rare disease exploration, the acquisition of domain knowledge remains a challenge. Traditional approaches like Reinforcement Learning from Human Feedback (RLHF) \citep{ouyang2022training,kaufmann2023survey} have demonstrated value in general settings, yet they struggle in contexts where expert knowledge is extremely expensive and sparse. This scenario is particularly pronounced in domains where domain expertise is fragmented among professionals with diverse competencies and availability constraints \citep{szymanski2025limitations,dhar2024enabling}. Consequently, there is a pressing need for novel approaches that can efficiently integrate domain expert feedback into LLMs while operating under tight budgetary and expert availability restrictions.

\begin{figure}[t]
\centering 
\includegraphics[width=\linewidth]{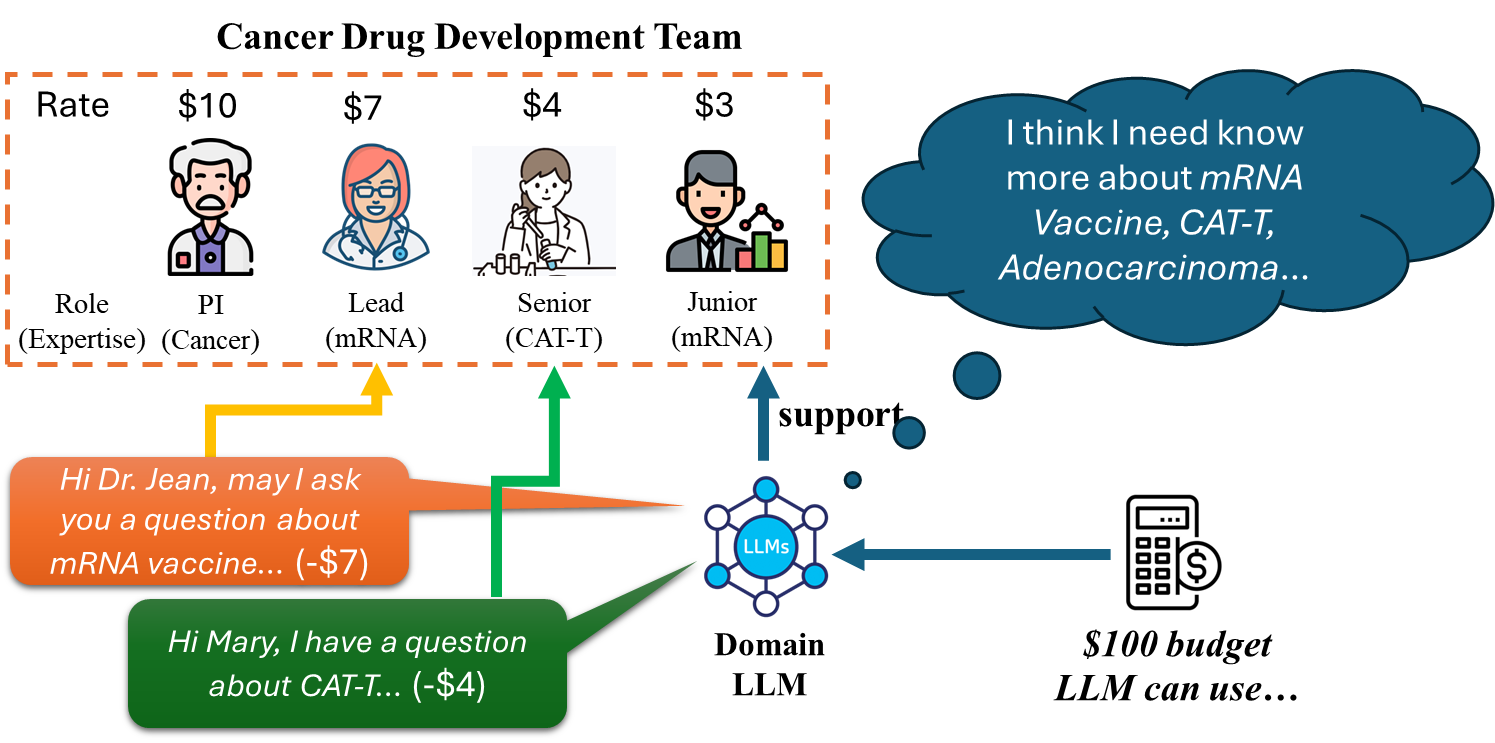}
\caption{Domain LLM Knowledge Acquisition via Cost-Efficient, Expert-Involved Interaction. The diagram depicts how PU-ADKA selectively engages domain experts with varying expertise and costs to acquire knowledge efficiently within a limited budget.} 
\label{fig:intro}
\end{figure}

To respond to this demand, we propose Positive Unlabeled Active Domain Knowledge Acquisition (PU-ADKA), which is designed to selectively engage with domain experts and acquire targeted feedback that can significantly enhance the performance of LLMs in specialized fields. Unlike conventional fine-tuning methods that passively incorporate affordable human feedback \citep{zhang2023llmaaa,wu2024knowledge}, PU-ADKA actively queries the most appropriate expert from a team given each member's computational profile. The model can elaborately consider factors such as the candidate expert's knowledge boundary, cost of consultation, and expert availability, thereby optimizing the knowledge acquisition process within a fixed budget (e.g., total \$100). The model training process leveraged newly released PubMed publications \citep{pubmed2024}, legacy architectures of LLMs and innovative simulations of expert-domain knowledge interactions. Through an intelligent knowledge selection process and cost-aware querying mechanism, PU-ADKA bridges the gap between the limited availability of expert input and the high demand for domain-specific information.

Figure \ref{fig:intro} illustrates the concept behind the proposed PU-ADKA. In this case, a domain LLM acknowledges gaps in its knowledge related to topics like \textit{mRNA vaccines}, \textit{CAT-T}, and \textit{adenocarcinoma} (to support a cancer drug development team) \citep{patel2025evoflow,yao2025intelligent}. Instead of relying on static, pre-existing datasets, PU-ADKA selectively engages with domain experts to acquire precise knowledge within a limited budget. The model evaluates the expertise, cost, and availability of different specialists, including PI, lead, senior, and junior scholars, to optimize knowledge acquisition. For example, in the image, the LLM selectively queries Dr. Jean for insights on \textit{mRNA vaccines} at a cost of \$7, while consulting Mary, a different expert, about \textit{CAT-T} for \$4, ensuring cost-effective expert engagement. This dynamic querying mechanism allows the LLM to refine its domain knowledge efficiently, making it particularly useful in critical domains like drug discovery and rare disease research, where expert knowledge is both sparse and expensive.

Our contributions in this paper are threefold and can be summarized as follows:

$\bullet$ We propose PU-ADKA, a cost-aware framework that strategically selects and queries domain experts by considering their availability, knowledge scope, and consultation cost, in order to enhance LLM performance under limited expert access and fixed budget constraints. \par
$\bullet$ We introduce the Cost-effective Knowledge Acquisition Dataset (CKAD), a new benchmark for LLM domain knowledge acquisition, to foster further research in the area of domain-specific LLM enhancement. \par
$\bullet$ We empirically validate the effectiveness of PU-ADKA through both simulation evaluation and a real-world cancer drug development study. The latter experiment involves a drug development team in which five experts with diverse backgrounds participate. The results show that PU-ADKA is promising in enhancing domain LLMs within a fixed budgetary restriction. \par

\section{Related Work}

\subsection{Human Feedback Integration in Domain-Specific LLMs} Domain-specific adaptation of LLMs has been advanced significantly by techniques such as domain-adaptive pretraining (DAPT) \citep{gururangan2020don} and various biomedical LLMs like BioMedLM \citep{bolton2024biomedlm}, ClinicalBLIP \citep{ji2024vision}, and BioGPT \citep{luo2022biogpt}. These methods effectively utilize large domain-specific corpora (e.g., PubMed) to incorporate static knowledge. However, they often fall short in capturing the dynamic insights from domain experts, crucial for rapidly evolving areas like drug discovery. RLHF~\citep{ouyang2022training} aims to align general LLMs with human preferences but typically depends on more homogeneous and less costly annotators, limiting its effectiveness in specialized domains where expert feedback is sparse and expensive. Attempts like ExpertQA \citep{malaviya2023expertqa} simulate multi-expert interactions but overlook practical constraints like budget limitations and asynchronous availability of experts. Our approach, \ours{}, overcomes these shortcomings by redefining expert knowledge acquisition as a budget-constrained optimization task, engaging experts based on their knowledge, cost, and availability, thereby transitioning from static data-driven adaptation to expert-guided learning.

\subsection{Budget-Constrained Active Learning with Multi-Expert Collaboration} Traditional active learning models primarily focus on maximizing sample information through uncertainty \citep{gal2017deep, kim2021lada,wang2024tag,yao2025elevating} or diversity \citep{chakraborty2015active, parvaneh2022active, citovsky2021batch}, often neglecting the varying costs associated with expert annotations, particularly in complex fields like biomedicine. Cost-sensitive approaches \citep{huang2017cost, henkel2023annotation,li2022batch} attempt to address this by optimizing for lower-cost annotators but fail to differentiate between the varied expertise levels necessary for accurately labeling complex cases. Unlike these methods, \ours{} integrates active learning with strategic expert collaboration, emphasizing both data sample selection based on the potential to update the model and efficient engagement of experts, balancing cost against their competency and availability.

\section{Methodology}

\subsection{Problem Definition}
\label{problem definition}
Given a fixed annotation budget \( B \), an unlabeled question pool 
\(
\mathcal{D}_{\text{tr}} = \{ q_i \}_{i=1}^{|\mathcal{D}_{\text{tr}}|}
\), and a team of domain experts \( \mathcal{E} = \{ e_j \}_{j=1}^{|\mathcal{E}|} \), our goal is to select an optimal set of ($q_i$, $e_j$) pairs to acquire expert-labeled data for  finetuning a large language model \( \theta \), maximizing finetuning performance on a target test set \( \mathcal{D}_{\text{te}} = \{ p_m \}_{m=1}^{|\mathcal{D}_{\text{te}}|} \).

Formally, we define an allocation function \( f: \mathcal{D}_{\text{tr}} \to \mathcal{E} \) that assigns each selected question \( q_i \) to an expert \( e_j \), ensuring that the total annotation cost remains within the budget \( B \). The optimization objective is:

 \begin{equation*}
\begin{gathered}
\mathcal{S}^* = \operatorname*{arg\,max }_{\mathcal{S} \subseteq \mathcal{D}_{\text{tr}} \times \mathcal{E}} \mathcal{F} (\theta_{\mathcal{S}}, \mathcal{D}_{\text{te}})\\
\text{s.t.,} \sum_{(q_i, e_j) \in \mathcal{S}} c(q_i, e_j) \leq B,
\end{gathered}
\end{equation*}
where, \( \mathcal{S}^* \) denotes the optimal set of ($q_i$, $e_j$) pairs that maximizes the performance metric \( \mathcal{F} (\theta_{\mathcal{S}}, \mathcal{D}_{\text{te}}) \) of the fine-tuned model \( \theta_{\mathcal{S}} \) on the target test set. The term \( c(q_i, e_j) \) represents the annotation cost incurred when expert \( e_j \) annotates question \( q_i \).

\begin{figure*}[t]
  \centering
  \includegraphics[width=1.0\linewidth]
  {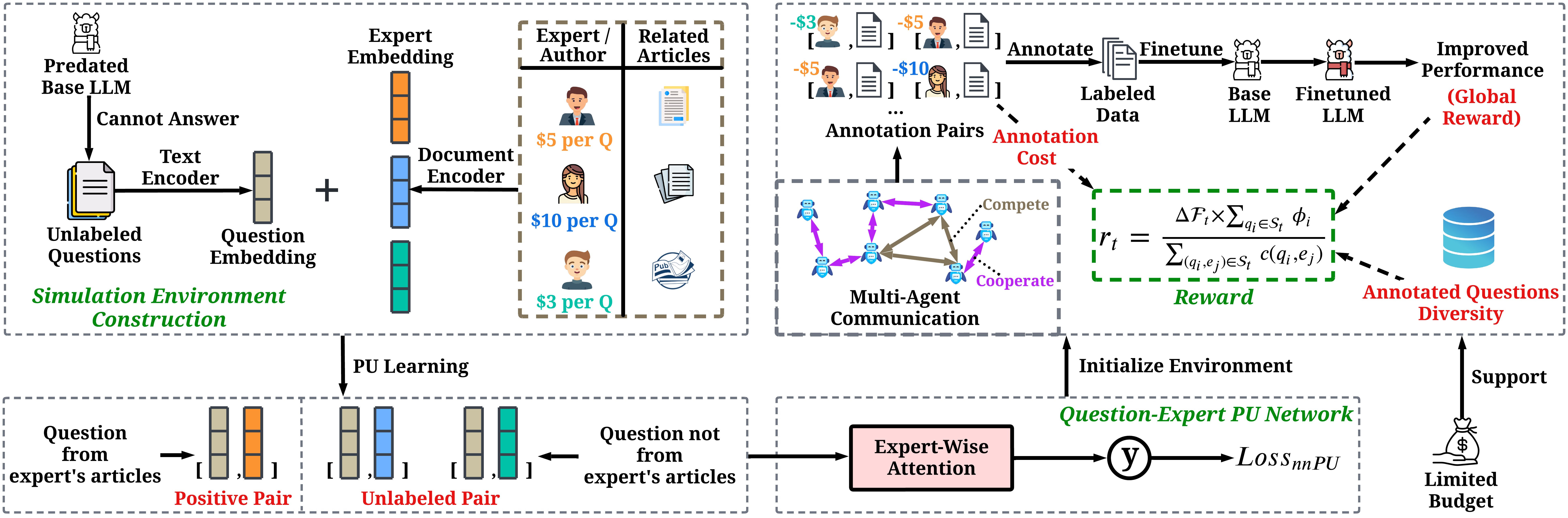}
  \caption{Illustration of our proposed PU-ADKA framework. Given an unlabeled question pool and a team of experts with varying expertise and cost, the question-expert PU learning network identifies the experts that can annotate specific questions based on limited positive examples. A multi-agent reinforcement learning module then selects which questions to annotate and assigns them to appropriate experts under a fixed budget. This process enables efficient acquisition of domain-specific knowledge for the base LLM through expert-in-the-loop supervision.}
  \label{framework}
\end{figure*}

\begin{table}[h]
\centering
\scriptsize
\caption{Notations.}
\label{tab:notations}
\setlength{\tabcolsep}{4pt}
\setlength{\extrarowheight}{0.1pt}
\begin{tabular}{|p{0.15\columnwidth}@{\hspace{2pt}}|p{0.78\columnwidth}@{\hspace{2pt}}|}
\hline
\textbf{Notation} & \textbf{Description} \\
\hline
\( B \) & Total annotation budget available. \\
\hline
\( D_{tr} \) & Unlabeled question pool for training. \\
\hline
\( D_{te} \) & Target test set for evaluation. \\
\hline
\( q_i \) & The \( i \)-th question in the unlabeled pool. \\
\hline
\( e_j \) & The \( j \)-th domain expert. \\
\hline
\( f \) & Allocation function assigning questions to experts. \\
\hline
\( \theta \) & Base language model. \\
\hline
\( \theta_S \) & Fine-tuned model using selected question-expert pairs \( S \). \\
\hline
\( S \) & Selected set of question–expert pairs for annotation. \\
\hline
\( c(q_i, e_j) \) & Cost for expert \( e_j \) to label question \( q_i \). \\
\hline
\( x_k^p \) & Positive question–expert pair used in PU learning. \\
\hline
\( x_k^u \) & Unlabeled question–expert pair used in PU learning. \\
\hline
\( \pi_p \) & Prior probability of a positive sample in PU learning. \\
\hline
\( g \) & Expert-wise attention network. \\
\hline
\( l(\cdot, \cdot) \) & Surrogate loss function (e.g., zero-one loss). \\
\hline
\( \Gamma^t_j \) & Number of times expert \( e_j \) has been selected up to time \( t \). \\
\hline
\( w^t_j \) & Sampling weight of expert \( e_j \) at time \( t \). \\
\hline
\( r_t \) & Reward at time step \( t \) in multi-agent RL. \\
\hline
\( \phi_i \) & Diversity score for question \( q_i \). \\
\hline
\( d(E^i_q, E^z_q) \) & Distance between question embeddings \( i \) and \( z \). \\
\hline
\( Z_i \) & Expert-wise representation of question \( q_i \). \\
\hline
\end{tabular}
\end{table}

\subsection{Simulation Environment Construction}
\label{simulation}
To facilitate our study, we introduce a novel benchmark dataset, CKAD, designed to simulate biomedical expert consultations and domain knowledge acquisition for LLMs. This dataset is constructed by strategically leveraging PubMed articles published after the knowledge cutoff date of the base model, ensuring that the selected content represents genuinely novel information. To further isolate new knowledge from prior model capabilities, we implement a temporal knowledge separation mechanism that enforces strict chronological boundaries between the base model’s existing knowledge and the newly acquired domain content. This is achieved through three key components detailed below:

\textbf{Predated Base Model Selection}: We employ Llama2-7B \citep{touvron2023llama} as our predated base model, chosen for its knowledge limitations to information available up to early 2023, prior to our target corpus. This temporal separation ensures a controlled setting for evaluating knowledge acquisition.

\textbf{Dataset Curation}: We construct CKAD from 2024 PubMed Central (PMC) \citep{pubmed2024}, extracting question-answer (QA) pairs using GPT-4o-2024-08-06 \citep{GPT-2024}. For each paper, five mechanism-focused QA pairs are generated using prompting\footnote{The detail of question-answer extraction prompt is provided in Appendix~\ref{prompt}.} and manually validated. To establish a well-isolated environment for assessing knowledge acquisition, we filter out QA pairs that can be answered by the base model. This process results in a final dataset of \textit{48,219 QA pairs} (the base model cannot correctly answer) representing post-2023 knowledge. To assess the quality of our dataset, we conduct a human evaluation on 100 randomly sampled QA pairs. Two PhD researchers with biomedical backgrounds independently scored each QA pair on a 1–5 scale\footnote{Quality scoring form is depicted in the Appendix~\ref{data_quality_rubric}}. The average score is 3.85, and Cohen’s Kappa \citep{mchugh2012interrater} between the evaluators is 0.73, reflecting high data quality and strong human agreement.

\textbf{Expert Simulation.} To simulate realistic annotation constraints, we construct a binary expert capability matrix \( A \in \mathbb{R}^{Q \times N} \), where \( A_{ji} = 1 \) indicates that expert \( e_j \) is assumed to be capable of annotating question \( q_i \), and 0 otherwise. This matrix is used to restrict which expert–question pairs are considered valid during simulation. Without such a constraint, every expert would be able to annotate every question, leading to minimal variation in annotation quality across experts—even for questions unrelated to their domain expertise. To construct \( A \), we use GPT-4o-2024-08-06 to analyze each expert’s publications and determine their capacity to annotate specific questions. The top 20 authors ranked by publication count are used as proxy experts. Each expert is assigned a per-question annotation rate, determined proportionally by the cumulative impact factor of their publications \citep{clarivate_mjl}.

\subsection{Positive Unlabeled Active Domain Knowledge Acquisition}
In this section, we present our Positive-Unlabeled Active Domain Knowledge Acquisition (PU-ADKA) framework, which selectively engages domain experts to acquire targeted feedback for improving LLM performance in specialized domains. PU-ADKA comprises two components: (1) Question–Expert Matching, formulated as a Positive-Unlabeled (PU) learning problem to model expert suitability; and (2) Multi-Agent Reinforcement Learning, which selects question–expert pairs under budget constraints. We elaborate on each component below.
\subsubsection{Expert Allocation with Positive Unlabeled Learning}
\label{qa matching}

\textbf{Motivation.} A key challenge in modeling expert–question suitability lies in the absence of explicit supervision: we can identify which expert authored the source publication from which a question is derived, and thus assume they are qualified to answer it; however, we cannot assume that all other experts are unqualified. This makes standard binary classification infeasible. To address this, we frame the question-expert matching task as a Positive–Unlabeled (PU) learning problem. Given a question–expert pair \((q_i, e_j)\), we label it as \textbf{\textit{positive}} if \(q_i\) originates from a publication authored by \(e_j\). If \(q_i\) does not come from \(e_j\)’s paper, we do not treat \((q_i, e_j)\) as a negative pair—instead, it remains \textbf{\textit{unlabeled}}, since the expert may still be qualified. For example, a scholar specializing in cancer NK cells may be able to answer a sepsis-related question involving extracellular vesicles, even without directly publishing in the sepsis domain.

\textbf{Model Training.}\quad We use LLM-based text representations, leveraging a pretrained Llama2-7B model to encode questions \( E_q^i \) and experts \( E_e^j \), with embeddings taken from the last hidden layer. Particularly, an expert's embedding is obtained by averaging the representations of their publications. To train our PU model to estimate expert knowledge boundary, we employ an expert-wise attention mechanism\footnote{The attention network is detailed in Appendix \ref{expert_atta}} $g$ and training with the non-negative PU risk estimator~\citep{kiryo2017positive}, which is defined as follows:
\begin{equation}
\small
    \begin{aligned}
        \text{Risk}_{pu}(g) = &\frac{\pi_p}{n_p} { \sum\limits_{i=1}^{n_p} } l(g(x_k^{p}),+1) + \\
    &max(0,\frac{1}{n_u} { \sum\limits_{i=1}^{n_u} } l(g(x_k^{u}),-1) - \\
    &\frac{\pi_p}{n_p} { \sum\limits_{i=1}^{n_p} } l(g(x_k^{p}),-1)),
    \label{nnpu}
    \end{aligned}
\end{equation}
where $\pi_p$ denotes positive class prior (\(\pi_p=0.1\) in our dataset), $l(\cdot, \cdot)$ is the surrogate loss of zero-one loss \citep{du2015convex,wu2023community}, $n_p$ represents the number of labeled positive instances, $n_u$ represents the number of unlabeled instances, $x_k^p$ and $x_k^u$ denote question-expert pairs in the labeled positive set and the unlabeled set, respectively.

\subsubsection{Domain Knowledge Acquisition via Multi-Agent Reinforcement Learning}
\label{multi agent rl}
The PU learning module is designed to estimate how well each expert aligns with a given question. This section builds on these estimates to select expert–question pairs for annotation under budget constraints, aiming to maximize domain knowledge acquisition.

\textbf{Motivation.} Effective knowledge acquisition requires selecting questions that are not only informative individually, but also complementary as a set. This necessitates modeling dependencies among questions—two high-value questions may become redundant when answered together. For example, questions about \textit{extracellular vesicles} in different disease contexts may overlap in the knowledge they elicit. Single-agent or greedy methods typically overlook such redundancy, leading to inefficient use of limited annotation budgets. To address this, we formulate the question selection as a multi-agent reinforcement learning (RL) problem, where each agent selects a question–expert pair while coordinating through shared rewards. This enables the model to account for inter-question dependencies and optimize the utility of the selected set.

\textbf{Multi-Agent RL State.}\quad The environment state is represented by a combination of features that capture both task-related and budgetary aspects: (1). The question--expert matching score \( g(q_i, e_j) \) is derived from the trained PU learning model and measures the suitability of assigning question \( q_i \) to expert \( e_j \). (2). The remaining budget \( B_t \) indicates the available annotation budget at time step \( t \). (3). The expert sampling weight $w_j^t$ quantifies the likelihood of selecting each expert \( e_j \), defined as:

\begin{equation}
    w_j^t = \frac{B_t}{c(q_i, e_j)} \times (1 - \alpha \Gamma_j^t),
\end{equation}
where \( \alpha \) is a decay factor, and \( \Gamma_j^t \) denotes the number of times expert \( e_j \) has been selected up to time step \( t \). This formulation encourages diversity in expert selection to enhance overall information gain while ensuring balanced workload distribution. 

\textbf{Multi-Agent Competition.} Different from previous studies, our framework allows multiple agents within the same model to simultaneously seek \((q_i, e_j)\) pairs, enabling different experts to compete for answering the same question. Leveraging our PU-based question-expert matching model, each question \( q_i \) is associated with a ranked list of potential experts. As a result, multiple experts \( e_1, e_2, \dots, e_h \) may select the same question \( q_i \). In such cases, \( q_i \) should be assigned to the expert with the highest matching score based on our PU matching network. To enforce this competitive selection, we introduce a competition function:

\begin{equation}
    \begin{aligned}
        Compete(&q_i \mid e_1, e_2, \dots, e_h) = e_v,  \\
        &\text{s.t.} \quad e_v = \operatorname*{arg\,max }_{e_j} g(q_i, e_j),
    \end{aligned}
\end{equation}
where \( g(q_i, e_j) \) represents the PU-based matching score between question \( q_i \) and expert \( e_j \), ensuring that the most suitable expert is selected. For experts who lose the competition for a given question in the current iteration,  the corresponding agents will then select alternative pairs and re-enter the competition process. This recursive procedure continues until all agents in the current state have been assigned unique questions.\par
\textbf{Multi-Agent Cooperation.}
To effectively encourage collaborative decision-making among agents and optimize knowledge acquisition under a fixed annotation budget, we define the reward function as:

\begin{equation}
    r_t = \frac{\Delta \mathcal{F}_t  \times \sum_{q_i \in \mathcal{S}_t} \phi_i}{\sum_{(q_i, e_j) \in \mathcal{S}_t} c(q_i, e_j)},
\end{equation}
where \( \Delta \mathcal{F}_t \) denotes the improvement in model performance on the validation set after incorporating newly labeled data at step t, and the denominator represents the total annotation cost~\citep{gao2020cost, huang2017cost, golazizian2024cost}. The diversity term \( \phi_i \) measures the distinctiveness of each selected question and is defined as:

\begin{equation}
    \phi_i = \min_{q_z \in \mathcal{S}_t} d(E_q^i, E_q^z),
\end{equation}
where \( \mathcal{S}_t \) denotes the current labeled question set, and \( d(\cdot, \cdot) \) is the Euclidean distance function. A larger \( \phi_i \) value indicates that the selected question is more diverse relative to past selections, thereby enhancing knowledge coverage and reducing redundancy.

\textbf{Model Training.}\quad To stabilize learning, we employ a Double DQN architecture~\citep{wang2020qplex}. The temporal-difference (TD) target $Y_t$ is computed as:

\begin{equation}
    Y_t = r_t + \gamma Q(s_{t+1}, \arg\max_{u_{t+1}} Q(s_{t+1}, u_{t+1}; \Omega_t); \Omega_t'),
\end{equation}
where \( s_{t+1} \) denotes the next state, \( \gamma \) is the discount factor, \( Q(s, u; \theta) \) is the action-value function, \( \Omega_t \) and \( \Omega_t' \) represent parameters of the policy and target network, respectively. To enhance generalization, we employ bootstrap sampling by selecting a random subset of experts (e.g., five per iteration) during the training stage. This strategy prevents overfitting to a specific set of experts, ensuring that the learned policy remains robust across diverse labeling scenarios.

\section{Experiments}

\subsection{Experimental Setup}
\label{experimental settings}

As described in Section~\ref{simulation}, we use the PubMed dataset for sepsis and cancer NK research from 2024 and adopt Llama2-7B as the base architecture. The experimental setup for our \ours{} model utilizes Llama2-7B with a sampling temperature of 1.0, a nucleus sampling top\_p value of 0.9, and a maximum token length of 4,096. The question and expert document encoders use the last hidden layer of Llama2-7B. For fine-tuning, we apply LoRA~\citep{hu2021lora} to improve training efficiency for large-scale models. The LoRA configuration includes a rank of 16, an alpha of 128, and a dropout rate of 0.1. Training involves learning LoRA matrices for all attention mechanisms in each configuration. The models are optimized using the AdamW optimizer with a learning rate of $2 \times 10^{-5}$. Each configuration undergoes three trials with different random seeds.

In the multi-agent reinforcement learning framework, we employ the Double DQN~\citep{wang2020qplex} architecture. The default number of agents is 10, with five experts selected per iteration. In each iteration, experts are ranked based on the sum of their papers' impact factors~\citep{clarivate_mjl}, and their unit prices are assigned accordingly as \([\$0.5, \$0.4, \$0.3, \$0.2, \$0.1]\) per labeled question. The total annotation budget is set to \$100. All implementations are conducted with Pytorch~\citep{paszke2017automatic}, PEFT~\citep{mangrulkar2022peft} and Transformers~\citep{wolf-etal-2020-transformers} on a computation node configured with a 64-core CPU and four 80GB H100 GPUs.

\subsection{Baselines} 
To ensure a comprehensive evaluation, our experiment includes a variety of baseline methodologies that encompass both question selection and expert allocation strategies. The comparison provides insights into the effectiveness of different active learning frameworks applied to LLMs. Below we detail the question selection used in baselines:\\ 
\textbf{RAND} - Questions are selected randomly, providing a baseline for minimal strategic intervention in data selection. \\
\textbf{DEITA} - \citet{liu2023makes} evaluates data across complexity, quality, and diversity using pretrained complexity scorer\footnote{https://huggingface.co/hkust-nlp/deita-complexity-scorer} and quality scorer \footnote{https://huggingface.co/hkust-nlp/deita-complexity-scorer} to score each unlabeled questions. \\
\textbf{CHERRY} - \citet{li2023quantity} applies the Instruction-Following Difficulty (IFD) metric to assess question quality autonomously. \\
\textbf{NUGGETS} - \citet{li2023one} assesses the relevance of questions by considering each as a single instance in one-shot learning contexts. \\
\textbf{LESS} -  \citet{xia2024less} calculates the influence of questions on the validation set to prioritize data that may yield the most significant insights during finetuning. \\
\textbf{ROSE} - \citet{wu2024rose} utilizes gradient similarity to evaluate the potential contribution of each question to the model’s performance.

For expert allocation, we implement the following methods:\\
\textbf{Random} - Experts are assigned randomly to questions. \\
\textbf{Cost-Greedy} - This method always selects the least expensive expert available, optimizing for cost efficiency. \\
\textbf{Match-Greedy} - Matches questions to experts based on the highest embedding similarity between them, facilitating a more informed allocation.

Each baseline represents a specific combination of question selection and expert allocation methods, providing a meaningful benchmark against which our proposed approach can be evaluated.

\begin{table*}[]
\caption{Overall performance comparison on CKAD dataset. The best result is highlighted in \textbf{bold}, and the second-best is \underline{underlined}. Numbers in parentheses denote the standard deviation across three runs.}
\centering
\renewcommand\arraystretch{0.97}
\label{tb:overall_result}
\scalebox{0.8}{
\begin{tabular}{l|l|c|c|c|c|c} 
\toprule
\multirow{2}{*}{\shortstack[l]{\textbf{Expert}\\\textbf{Allocation}}} & \multirow{2}{*}{\shortstack[l]{\textbf{Question}\\\textbf{Selection}}} & \bf GPT-4o-2024-08-06 & \bf GPT-4-Turbo & \bf GPT-4o-2024-08-06 & \bf GPT-4-Turbo & \bf Avg.Length \\ 
\cmidrule(lr){3-4} \cmidrule(lr){5-6}\cmidrule(lr){7-7}
& & WR (\%)& WR (\%)& LC\_WR  (\%)& LC\_WR (\%)& - \\ \midrule

\multirow[t]{6}{*}{Random} & RAND  & 4.7 \std{0.4} & 6.7 \std{0.8} & 20.3 \std{0.9} & 20.4 \std{0.8} & 2220 \\ 
& DEITA  & 9.6 \std{0.3} & 7.9 \std{0.1} & 21.0 \std{0.9} & 22.1 \std{0.8} & 2212 \\
& CHERRY  & 7.8 \std{0.1} & 8.3 \std{0.2} & 20.4 \std{0.9} & 21.5 \std{0.9} & 2221 \\
& NUGGETS  &  10.4 \std{0.1} & 10.7 \std{0.4} & 21.0 \std{0.8} & 20.4 \std{0.8} & 2204 \\
& LESS  & 7.9 \std{0.2} & 7.9 \std{0.2} & 22.0 \std{1.0} & \underline{24.0} \std{1.1} & 2212 \\
& ROSE  & 8.1 \std{0.4} & 10.0 \std{0.2} & 21.5 \std{1.0} & 22.7 \std{1.0} & 2194 \\ \midrule

\multirow[t]{6}{*}{Cost-Greedy} & RAND  & 6.2 \std{0.4} & 6.7 \std{0.8} & 20.4 \std{0.9} & 20.5 \std{0.9} & 2207 \\ 
& DEITA  & \underline{14.2} \std{0.8} & \underline{11.7} \std{0.2} & 20.9 \std{1.0} & 20.9 \std{0.9} & 2246  \\
& CHERRY  & 11.7 \std{0.3} & 10.0 \std{0.4} & 23.4 \std{0.9} & 22.1 \std{1.1} & 2236 \\
& NUGGETS  & 7.9 \std{0.4} & 8.7 \std{0.4} & 21.5 \std{0.9} & 20.4 \std{0.9} & 2182  \\
& LESS  & 12.1 \std{0.4} & 9.6 \std{0.4} & 22.1 \std{0.8} & 21.2 \std{1.0} & 2218  \\
& ROSE  & 8.3 \std{0.8} & 9.7 \std{0.2} & 20.4 \std{0.9} & 22.7 \std{1.0} & 2174 \\ \midrule

\multirow[t]{6}{*}{Match-Greedy} & RAND  & 6.7 \std{0.8} & 7.9 \std{0.4} & 20.9 \std{1.0} & 19.9 \std{0.8} & 2204 \\ 
& DEITA  & 10.0 \std{0.3} & 9.2 \std{0.8} & 21.2 \std{1.0} & 22.3 \std{0.9} & 2214 \\
& CHERRY  & 7.5 \std{0.0} & 9.2 \std{0.2} & 21.0 \std{0.9} & 23.3 \std{1.1} & 2173 \\
& NUGGETS  & 9.5 \std{0.3} & 11.6 \std{0.2} & 22.1 \std{1.0} & 21.6 \std{0.9} & 2182 \\
& LESS  & 12.1 \std{0.4} & 10.4 \std{0.2} & \underline{23.5} \std{1.0} & 22.5 \std{1.0} & 2252 \\
& ROSE  & 9.2 \std{0.1} & 10.9 \std{0.4} & 22.5 \std{0.9} & 21.9 \std{1.0} & 2229 \\ \midrule


\rowcolor{violet!15} \multirow{1}{*}{Ours} & \ours{} & \textbf{18.2} \std{0.6} & \textbf{16.7} \std{0.4} & \textbf{25.6} \std{1.0} & \textbf{26.5} \std{0.9} & 1781 \\ \bottomrule
\end{tabular}
} 
\end{table*}

\subsection{Evaluation Benchmarks and Metrics.} 
To ensure a clean evaluation of knowledge acquisition, our CKAD dataset consists of general disease mechanism question-answer pairs\footnote{Dataset statistics are provided in Appendix~\ref{appendix:data statictics}.} that cannot be answered by base LLM (Llama2-7B) initially (i.e., the initial answerable rate is 0). During the simulation training stage, we employ two advanced models, GPT-4o-2024-08-06 \citep{GPT-2024} and GPT-4-Turbo \citep{achiam2023gpt}, as judge models. The evaluation metrics include: \textbf{Win Rate} (WR), which measures the percentage of instances where the judge LLM determines that the model-generated answer adequately captures the core meaning of the golden answer; and \textbf{Length-Controlled Win Rate} (LC\_WR) \citep{dubois2404length}, a variant of WR that filters out samples with large answer length discrepancies between the model output and the golden answer, helping to control for verbosity bias during evaluation.

Additionally, following \citet{wang2023chain}, we conduct human-involved experiments to validate the effectiveness of our method. The expert team consists of three sepsis specialists and two cancer specialists, representing different levels of expertise. Among them, one expert is a medical doctor, and the remaining four are PhD students\footnote{Detailed information about the human experts is provided in \nameref{ethics}.}.

\subsection{Experimental Results}
Our experimental results are detailed in Table~\ref{tb:overall_result}, where we compare the performance of our method, \ours{}, against various baseline strategies. \ours{} consistently outperforms all baselines in terms of knowledge acquisition across different judging models. Specifically, with the GPT-40-2024-08-06 model as judge, \ours{} achieves a WR of 18.2\% and an LC\_WR of 25.65\%. When evaluated by the GPT-4-Turbo model, it records a WR of 16.7\% and an LC\_WR of 26.57\%. These results exceed those of the next best baseline, DEITA under the Cost-Greedy strategy, by margins of 4\% and 5\% in WR, and 2.1\% and 3.2\% in LC\_WR, respectively, under the two judging conditions. Noteably, LESS performs stable when under both Cost-Greedy and Match-Greedy settings, the GPT-4o-2024-08-06 and GPT-4-Turbo judge the WR at 12.1\% and 10\% in both settings. Furthermore, the minimal baseline performance under fully random conditions, with WR of 4.7\% and 6.7\%, highlights the baseline challenge and emphasizes the robustness of our method against less strategic approaches.
\begin{table}[htp]  
  \centering
  \caption{Human-involved results judged by GPT-4-Turbo.}
  \setlength{\tabcolsep}{1.8mm} 
  \resizebox{\linewidth}{!}{%
    \begin{tabular}{lcc}
    \toprule
    & WR (\%)& LC\_WR (\%)\\ \midrule
    Random (Random) & 7.5 \std{0.7} & 20.3 \std{0.8} \\ 
    LESS (Random)   & 9.2 \std{0.5} & 20.5\std{0.9} \\ 
    LESS (Cost-Greedy)  & 11.4 \std{0.6} & 21.0 \std{1.0} \\ 
    LESS (Match-Greedy) & \underline{12.5} \std{0.7} & \underline{21.2} \std{0.8} \\ 
    \ours{} & \textbf{15.2} \std{0.8} & \textbf{24.3} \std{0.9} \\ 
    \bottomrule
    \end{tabular}
    }
  \label{human}
\end{table}

\subsection{Human Involved Validation}
To further substantiate the robustness of our method, \ours{}, we implement it within a professional biomedical team of experts under a simulated budget constraint of \$100 per game. The cost of human experts is varied, reflecting their respective professional knowledge in the domain, with unit prices set at [\$0.5, \$0.2, \$0.1, \$0.1, \$0.1] per labeled question. We assess the performance in terms of WR and LC\_WR using GPT-4-Turbo as the judge under various settings: fully random, and LESS for question selection combined with each of the three expert allocation strategies (Random, Cost-Greedy, and Match-Greedy). The detailed results are presented in Table~\ref{human}.
\begin{figure}[h]
\centering 
\includegraphics[width=\linewidth]{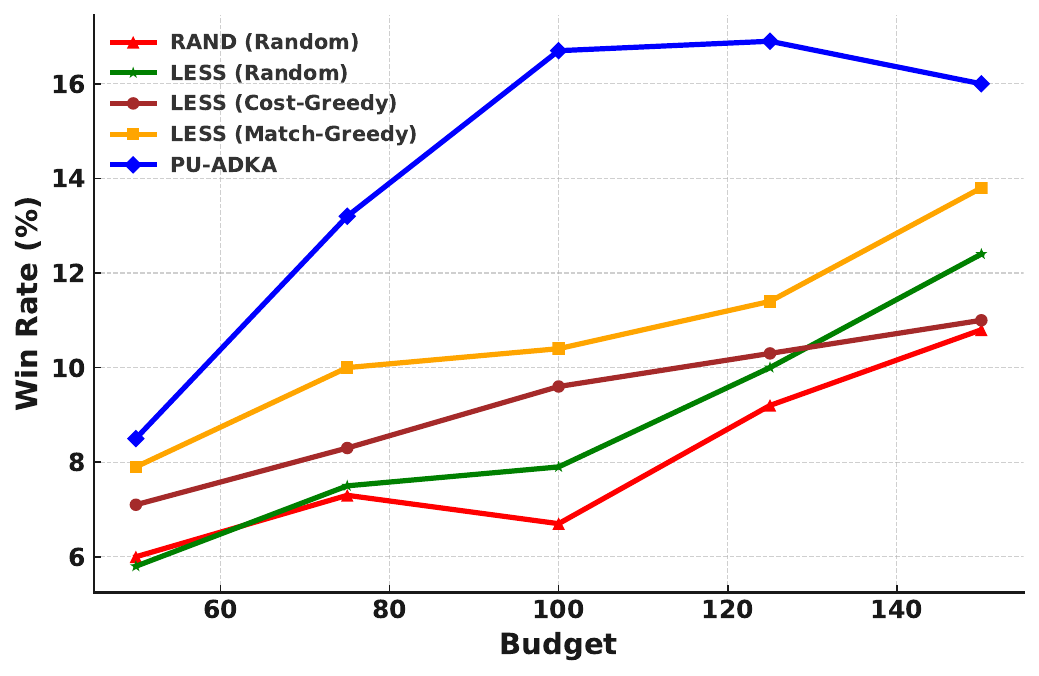}
\caption{Performance comparison under different budgets (\$) evaluated by GPT-4-Turbo} 
\label{budget_plot}
\end{figure}
\begin{table}[htp]  
  \centering
  \caption{Ablation results on CKAD dataset with \cmark\ indicating the enabling of the corresponding module. Evaluation performed by GPT-4-Turbo.}
  \setlength{\tabcolsep}{1.8mm} 
  \resizebox{\linewidth}{!}{%
  \begin{tabular}{c|cc|cc}
    \hline
    Variant & PU & MA & WR (\%) & LC\_WR (\%)\\
    \hline
    I     & \xmark & \cmark  & 13.3 \std{0.7} &  \underline{23.2} \std{1.1} \\
    II    & \cmark & \xmark  & \underline{14.2} \std{0.6} &  23.0 \std{1.0} \\
    \hline
    \ours{} & \cmark & \cmark & \textbf{16.7} \std{0.4} & \textbf{26.5} \std{0.9}  \\
    \hline
  \end{tabular}
  }
  \label{tab:ablation}
\end{table}

The results reveal that \ours{} notably surpasses the most competitive baseline, LESS (Match-Greedy), by margins of 2.7\% and 3.1\% in WR and LC\_WR, respectively. This enhancement in performance in a practical setting underscores the effectiveness of our method, particularly in scenarios constrained by budget. This real-world application not only validates the utility of \ours{} but also establishes it as a formidable approach in the domain of budget-limited active learning.

\subsection{Ablation Study}
\subsubsection{Validating the Utility of Each Module} To thoroughly assess the contributions of each component within PU-ADKA, specifically the multi-agent (MA) framework and the positive-unlabeled (PU) learning approach, we perform a series of ablation studies. These studies are conducted on the QA dataset, with GPT-4-Turbo serving as the judge. We explore two key variants:\\ \textbf{Variant I} - Utilizes unsupervised embedding-based similarity measures in place of the PU learning model to understand the impact of the PU approach on the overall performance. \\
\textbf{Variant II} - Operates under a single-agent setup to evaluate the effectiveness of our multi-agent configuration.

The results, detailed in Table~\ref{tab:ablation}, highlight the integral role each module plays in the success of PU-ADKA. The comparison with Variant I underscores the superiority of our PU-based question-expert matching technique. Similarly, when contrasted with the single-agent model of Variant II, our multi-agent method demonstrates its enhanced capability in expert allocation strategy, confirming the benefits of our comprehensive framework in active learning scenarios.

\subsubsection{Performance under Different Budgets} Following \citet{hacohen2022active,li2022batch}, we evaluate the performance of our model, PU-ADKA, against various baseline methods under differing budget scenarios, as depicted in Figure~\ref{budget_plot}. The results indicate that our method achieves consistently robust outcomes across all tested budget levels compared to the baselines. Notably, at a budget of \$100, PU-ADKA significantly outperforms the next best approach, LESS (Match-Greedy). Beyond this budget point, the rate of knowledge acquisition stabilizes, showing no substantial further increases. This observation suggests that our method is particularly effective at rapidly acquiring knowledge within constrained budget settings, demonstrating a distinct advantage over competing methods in efficiently utilizing available resources.

\section{Conclusion and Future Work}

We propose PU-ADKA, a cost-aware active learning framework that enhances LLMs by selectively engaging domain experts based on their expertise, availability, and annotation cost. PU-ADKA improves budget efficiency and LLM performance, as validated through simulations and real-world biomedical tasks. The release of the CKAD dataset supports further research on domain-specific LLM tuning. In future work, we plan to explore alternative backbone models and extend PU-ADKA to specialized domains beyond biomedical domain.
\clearpage

\section*{Limitations}
$\bullet$ \textit{Scalability with Increasing Data and Experts.} As the number of unlabeled data points and available experts grows, the scale of PU-ADKA changes significantly. Larger datasets require more efficient selection strategies, while an increasing pool of experts introduces greater complexity in allocation and coordination. Future research should explore more scalable solutions to maintain efficiency as the system scales to real-world, large-scale applications.\\ $\bullet$ \textit{Impact of Number of Agents and Computational Constraints.}  
The number of agents directly affects the system’s performance and computational demands. While PU-ADKA operates within a multi-agent framework, we do not extensively experiment with varying agent numbers due to the high computational cost associated with training and coordination. Additionally, we do not explore different batch sizes or report computational efficiency under varying agent settings. Future work should investigate the trade-offs between agent scalability, computational efficiency, and performance optimization.\\ 
$\bullet$ \textit{Generalizability to Other Domains.}  
While this study primarily focuses on biomedical expert interactions, other high-cost domains such as law and finance face similar challenges. Expanding PU-ADKA to these fields and evaluating its adaptability to different datasets and model architectures will be essential for broader applicability.\\
$\bullet$ \textit{Backbone Diversity and Model Size.}
We adopt Llama2-7B as the fixed backbone to ensure consistent evaluation. However, this limits the exploration of PU-ADKA's effectiveness across other model families and sizes. Future work should investigate its adaptability to diverse architectures, including larger models, instruction-tuned variants, and domain-specific LLMs.

\section*{Ethics Statement}
\addcontentsline{toc}{section}{Ethics Statement}
\phantomsection
\label{ethics}

All human annotation work in this study is conducted by domain experts in an external biomedical research group. Importantly, none of the experts are listed as co-authors of this paper. The process is coordinated by a biomedical Principal Investigator (PI), who is a co-author, but does not participate in any annotation work directly. The experts include one medical doctor and four PhD-level biomedical researchers, and they are blind to the study’s hypotheses, model design, and experiments.

All annotation work is performed during the experts’ regular paid working hours as part of their institutional responsibilities, and no additional compensation is provided. To simulate annotation cost in our experiments, we adopt a relative cost scheme based on typical salary ratios across seniority levels (e.g., doctor : senior PhD : junior PhD = 5 : 2 : 1), without disclosing any actual salary details. All experts agree that their annotations will be used in this study and released alongside the dataset. The content they annotate is derived entirely from publicly available biomedical literature and contains no personal or sensitive information. Accordingly, no additional ethics board review is required under the ACL Ethics Policy.

\bibliography{anthology,custom}

\begin{thebibliography}{51}
\expandafter\ifx\csname natexlab\endcsname\relax\def\natexlab#1{#1}\fi

\bibitem[{Achiam et~al.(2023)Achiam, Adler, Agarwal, Ahmad, Akkaya, Aleman, Almeida, Altenschmidt, Altman, Anadkat et~al.}]{achiam2023gpt}
Josh Achiam, Steven Adler, Sandhini Agarwal, Lama Ahmad, Ilge Akkaya, Florencia~Leoni Aleman, Diogo Almeida, Janko Altenschmidt, Sam Altman, Shyamal Anadkat, et~al. 2023.
\newblock Gpt-4 technical report.
\newblock \emph{arXiv preprint arXiv:2303.08774}.

\bibitem[{Bolton et~al.(2024)Bolton, Venigalla, Yasunaga, Hall, Xiong, Lee, Daneshjou, Frankle, Liang, Carbin et~al.}]{bolton2024biomedlm}
Elliot Bolton, Abhinav Venigalla, Michihiro Yasunaga, David Hall, Betty Xiong, Tony Lee, Roxana Daneshjou, Jonathan Frankle, Percy Liang, Michael Carbin, et~al. 2024.
\newblock Biomedlm: A 2.7 b parameter language model trained on biomedical text.
\newblock \emph{arXiv preprint arXiv:2403.18421}.

\bibitem[{Chakraborty et~al.(2015)Chakraborty, Balasubramanian, Sun, Panchanathan, and Ye}]{chakraborty2015active}
Shayok Chakraborty, Vineeth Balasubramanian, Qian Sun, Sethuraman Panchanathan, and Jieping Ye. 2015.
\newblock Active batch selection via convex relaxations with guaranteed solution bounds.
\newblock \emph{IEEE transactions on pattern analysis and machine intelligence}, 37(10):1945--1958.

\bibitem[{Citovsky et~al.(2021)Citovsky, DeSalvo, Gentile, Karydas, Rajagopalan, Rostamizadeh, and Kumar}]{citovsky2021batch}
Gui Citovsky, Giulia DeSalvo, Claudio Gentile, Lazaros Karydas, Anand Rajagopalan, Afshin Rostamizadeh, and Sanjiv Kumar. 2021.
\newblock Batch active learning at scale.
\newblock \emph{Advances in Neural Information Processing Systems}, 34:11933--11944.

\bibitem[{{Clarivate}(2025)}]{clarivate_mjl}
{Clarivate}. 2025.
\newblock \href {https://mjl.clarivate.com/home} {Master journal list}.
\newblock Accessed: 2025-01-02.

\bibitem[{Devlin et~al.(2019)Devlin, Chang, Lee, and Toutanova}]{devlin2019bert}
Jacob Devlin, Ming-Wei Chang, Kenton Lee, and Kristina Toutanova. 2019.
\newblock Bert: Pre-training of deep bidirectional transformers for language understanding.
\newblock In \emph{Proceedings of the 2019 conference of the North American chapter of the association for computational linguistics: human language technologies, volume 1 (long and short papers)}, pages 4171--4186.

\bibitem[{Dhar(2024)}]{dhar2024enabling}
UU~Dhar. 2024.
\newblock Enabling domain expert evaluation of emerging ai technologies in healthcare settings.

\bibitem[{Du~Plessis et~al.(2015)Du~Plessis, Niu, and Sugiyama}]{du2015convex}
Marthinus Du~Plessis, Gang Niu, and Masashi Sugiyama. 2015.
\newblock Convex formulation for learning from positive and unlabeled data.
\newblock In \emph{International conference on machine learning}, pages 1386--1394. PMLR.

\bibitem[{Dubois et~al.(2024{\natexlab{a}})Dubois, Galambosi, Liang, and Hashimoto}]{dubois2404length}
Yann Dubois, Bal{\'a}zs Galambosi, Percy Liang, and Tatsunori~B Hashimoto. 2024{\natexlab{a}}.
\newblock Length-controlled alpacaeval: A simple way to debias automatic evaluators,.
\newblock \emph{URL https://arxiv. org/abs/2404.04475}.

\bibitem[{Dubois et~al.(2024{\natexlab{b}})Dubois, Galambosi, Liang, and Hashimoto}]{dubois2024length}
Yann Dubois, Bal{\'a}zs Galambosi, Percy Liang, and Tatsunori~B Hashimoto. 2024{\natexlab{b}}.
\newblock Length-controlled alpacaeval: A simple way to debias automatic evaluators.
\newblock \emph{arXiv preprint arXiv:2404.04475}.

\bibitem[{Gal et~al.(2017)Gal, Islam, and Ghahramani}]{gal2017deep}
Yarin Gal, Riashat Islam, and Zoubin Ghahramani. 2017.
\newblock Deep bayesian active learning with image data.
\newblock In \emph{International conference on machine learning}, pages 1183--1192. PMLR.

\bibitem[{Gao and Saar-Tsechansky(2020)}]{gao2020cost}
Ruijiang Gao and Maytal Saar-Tsechansky. 2020.
\newblock Cost-accuracy aware adaptive labeling for active learning.
\newblock In \emph{Proceedings of the AAAI conference on artificial intelligence}, volume~34, pages 2569--2576.

\bibitem[{Golazizian et~al.(2024)Golazizian, Ziabari, Omrani, and Dehghani}]{golazizian2024cost}
Preni Golazizian, Alireza~S Ziabari, Ali Omrani, and Morteza Dehghani. 2024.
\newblock Cost-efficient subjective task annotation and modeling through few-shot annotator adaptation.
\newblock \emph{arXiv preprint arXiv:2402.14101}.

\bibitem[{Gururangan et~al.(2020)Gururangan, Marasovi{\'c}, Swayamdipta, Lo, Beltagy, Downey, and Smith}]{gururangan2020don}
Suchin Gururangan, Ana Marasovi{\'c}, Swabha Swayamdipta, Kyle Lo, Iz~Beltagy, Doug Downey, and Noah~A Smith. 2020.
\newblock Don't stop pretraining: Adapt language models to domains and tasks.
\newblock \emph{arXiv preprint arXiv:2004.10964}.

\bibitem[{Hacohen et~al.(2022)Hacohen, Dekel, and Weinshall}]{hacohen2022active}
Guy Hacohen, Avihu Dekel, and Daphna Weinshall. 2022.
\newblock Active learning on a budget: Opposite strategies suit high and low budgets.
\newblock \emph{arXiv preprint arXiv:2202.02794}.

\bibitem[{Henkel et~al.(2023)Henkel, Hoxha, Sumbul, M{\"o}llenbrok, and Demir}]{henkel2023annotation}
Julia Henkel, Genc Hoxha, Gencer Sumbul, Lars M{\"o}llenbrok, and Beg{\"u}m Demir. 2023.
\newblock Annotation cost efficient active learning for content based image retrieval.
\newblock In \emph{IGARSS 2023-2023 IEEE International Geoscience and Remote Sensing Symposium}, pages 4994--4997. IEEE.

\bibitem[{Hu et~al.(2021)Hu, Shen, Wallis, Allen-Zhu, Li, Wang, Wang, and Chen}]{hu2021lora}
Edward~J Hu, Yelong Shen, Phillip Wallis, Zeyuan Allen-Zhu, Yuanzhi Li, Shean Wang, Lu~Wang, and Weizhu Chen. 2021.
\newblock Lora: Low-rank adaptation of large language models.
\newblock \emph{arXiv preprint arXiv:2106.09685}.

\bibitem[{Huang et~al.(2017)Huang, Chen, Mu, and Zhou}]{huang2017cost}
Sheng-Jun Huang, Jia-Lve Chen, Xin Mu, and Zhi-Hua Zhou. 2017.
\newblock Cost-effective active learning from diverse labelers.
\newblock In \emph{IJCAI}, pages 1879--1885.

\bibitem[{Ji et~al.(2024)Ji, Hou, Chen, Pan, and Xiang}]{ji2024vision}
Jia Ji, Yongshuai Hou, Xinyu Chen, Youcheng Pan, and Yang Xiang. 2024.
\newblock Vision-language model for generating textual descriptions from clinical images: model development and validation study.
\newblock \emph{JMIR Formative Research}, 8:e32690.

\bibitem[{Kaufmann et~al.(2023)Kaufmann, Weng, Bengs, and H{\"u}llermeier}]{kaufmann2023survey}
Timo Kaufmann, Paul Weng, Viktor Bengs, and Eyke H{\"u}llermeier. 2023.
\newblock A survey of reinforcement learning from human feedback.
\newblock \emph{arXiv preprint arXiv:2312.14925}.

\bibitem[{Kim et~al.(2021)Kim, Song, Jang, and Moon}]{kim2021lada}
Yoon-Yeong Kim, Kyungwoo Song, JoonHo Jang, and Il-Chul Moon. 2021.
\newblock Lada: Look-ahead data acquisition via augmentation for deep active learning.
\newblock \emph{Advances in Neural Information Processing Systems}, 34:22919--22930.

\bibitem[{Kiryo et~al.(2017)Kiryo, Niu, Du~Plessis, and Sugiyama}]{kiryo2017positive}
Ryuichi Kiryo, Gang Niu, Marthinus~C Du~Plessis, and Masashi Sugiyama. 2017.
\newblock Positive-unlabeled learning with non-negative risk estimator.
\newblock \emph{Advances in neural information processing systems}, 30.

\bibitem[{Li et~al.(2023{\natexlab{a}})Li, Zhang, Li, Chen, Chen, Cheng, Wang, Zhou, and Xiao}]{li2023quantity}
Ming Li, Yong Zhang, Zhitao Li, Jiuhai Chen, Lichang Chen, Ning Cheng, Jianzong Wang, Tianyi Zhou, and Jing Xiao. 2023{\natexlab{a}}.
\newblock From quantity to quality: Boosting llm performance with self-guided data selection for instruction tuning.
\newblock \emph{arXiv preprint arXiv:2308.12032}.

\bibitem[{Li et~al.(2022)Li, Phillips, Yu, Kirby, and Zhe}]{li2022batch}
Shibo Li, Jeff~M Phillips, Xin Yu, Robert Kirby, and Shandian Zhe. 2022.
\newblock Batch multi-fidelity active learning with budget constraints.
\newblock \emph{Advances in Neural Information Processing Systems}, 35:995--1007.

\bibitem[{Li et~al.(2023{\natexlab{b}})Li, Hui, Xia, Yang, Yang, Zhang, Si, Liu, Liu, Huang et~al.}]{li2023one}
Yunshui Li, Binyuan Hui, Xiaobo Xia, Jiaxi Yang, Min Yang, Lei Zhang, Shuzheng Si, Junhao Liu, Tongliang Liu, Fei Huang, et~al. 2023{\natexlab{b}}.
\newblock One shot learning as instruction data prospector for large language models.
\newblock \emph{arXiv preprint arXiv:2312.10302}.

\bibitem[{Liu et~al.(2023)Liu, Zeng, He, Jiang, and He}]{liu2023makes}
Wei Liu, Weihao Zeng, Keqing He, Yong Jiang, and Junxian He. 2023.
\newblock What makes good data for alignment? a comprehensive study of automatic data selection in instruction tuning.
\newblock \emph{arXiv preprint arXiv:2312.15685}.

\bibitem[{Luo et~al.(2022)Luo, Sun, Xia, Qin, Zhang, Poon, and Liu}]{luo2022biogpt}
Renqian Luo, Liai Sun, Yingce Xia, Tao Qin, Sheng Zhang, Hoifung Poon, and Tie-Yan Liu. 2022.
\newblock Biogpt: generative pre-trained transformer for biomedical text generation and mining.
\newblock \emph{Briefings in bioinformatics}, 23(6):bbac409.

\bibitem[{Malaviya et~al.(2023)Malaviya, Lee, Chen, Sieber, Yatskar, and Roth}]{malaviya2023expertqa}
Chaitanya Malaviya, Subin Lee, Sihao Chen, Elizabeth Sieber, Mark Yatskar, and Dan Roth. 2023.
\newblock Expertqa: Expert-curated questions and attributed answers.
\newblock \emph{arXiv preprint arXiv:2309.07852}.

\bibitem[{Mangrulkar et~al.(2022)Mangrulkar, Somasundaram, and Shrivastava}]{mangrulkar2022peft}
Sanket Mangrulkar, Kaustubh Somasundaram, and Akhilesh Shrivastava. 2022.
\newblock Peft: Parameter-efficient fine-tuning.
\newblock \emph{Hugging Face}.
\newblock \url{https://github.com/huggingface/peft}.

\bibitem[{McHugh(2012)}]{mchugh2012interrater}
Mary~L McHugh. 2012.
\newblock Interrater reliability: the kappa statistic.
\newblock \emph{Biochemia medica}, 22(3):276--282.

\bibitem[{Naveed et~al.(2023)Naveed, Khan, Qiu, Saqib, Anwar, Usman, Akhtar, Barnes, and Mian}]{naveed2023comprehensive}
Humza Naveed, Asad~Ullah Khan, Shi Qiu, Muhammad Saqib, Saeed Anwar, Muhammad Usman, Naveed Akhtar, Nick Barnes, and Ajmal Mian. 2023.
\newblock A comprehensive overview of large language models.
\newblock \emph{arXiv preprint arXiv:2307.06435}.

\bibitem[{{OpenAI}(2024)}]{GPT-2024}
{OpenAI}. 2024.
\newblock \href {https://openai.com/index/hello-gpt-4o/} {Gpt-4o model card}.

\bibitem[{Ouyang et~al.(2022)Ouyang, Wu, Jiang, Almeida, Wainwright, Mishkin, Zhang, Agarwal, Slama, Ray et~al.}]{ouyang2022training}
Long Ouyang, Jeffrey Wu, Xu~Jiang, Diogo Almeida, Carroll Wainwright, Pamela Mishkin, Chong Zhang, Sandhini Agarwal, Katarina Slama, Alex Ray, et~al. 2022.
\newblock Training language models to follow instructions with human feedback.
\newblock \emph{Advances in neural information processing systems}, 35:27730--27744.

\bibitem[{Pal et~al.(2024)Pal, Bhattacharya, Lee, and Chakraborty}]{pal2024domain}
Soumen Pal, Manojit Bhattacharya, Sang-Soo Lee, and Chiranjib Chakraborty. 2024.
\newblock A domain-specific next-generation large language model (llm) or chatgpt is required for biomedical engineering and research.
\newblock \emph{Annals of biomedical engineering}, 52(3):451--454.

\bibitem[{Parvaneh et~al.(2022)Parvaneh, Abbasnejad, Teney, Haffari, Van Den~Hengel, and Shi}]{parvaneh2022active}
Amin Parvaneh, Ehsan Abbasnejad, Damien Teney, Gholamreza~Reza Haffari, Anton Van Den~Hengel, and Javen~Qinfeng Shi. 2022.
\newblock Active learning by feature mixing.
\newblock In \emph{Proceedings of the IEEE/CVF conference on computer vision and pattern recognition}, pages 12237--12246.

\bibitem[{Paszke et~al.(2017)Paszke, Gross, Massa, Lerer, Bradbury, Chanan, Killeen, Lin, Gimelshein, Antiga et~al.}]{paszke2017automatic}
Adam Paszke, Sam Gross, Francisco Massa, Adam Lerer, James Bradbury, Gregory Chanan, Trevor Killeen, Zeming Lin, Natalia Gimelshein, Luca Antiga, et~al. 2017.
\newblock Automatic differentiation in pytorch.
\newblock In \emph{Advances in Neural Information Processing Systems}.

\bibitem[{Patel et~al.(2025)Patel, Peng, Fraser, Friedman, Chatterjee, and Yao}]{patel2025evoflow}
Sawan Patel, Fred~Zhangzhi Peng, Keith Fraser, Adam~D Friedman, Pranam Chatterjee, and Sherwood Yao. 2025.
\newblock Evoflow-rna: Generating and representing non-coding rna with a language model.
\newblock \emph{bioRxiv}, pages 2025--02.

\bibitem[{{PubMed}(2024)}]{pubmed2024}
{PubMed}. 2024.
\newblock {PubMed Data: Download and Use}.
\newblock \url{https://pubmed.ncbi.nlm.nih.gov/download}.
\newblock Last updated Dec 14, 2024.

\bibitem[{Szymanski et~al.(2025)Szymanski, Ziems, Eicher-Miller, Li, Jiang, and Metoyer}]{szymanski2025limitations}
Annalisa Szymanski, Noah Ziems, Heather~A Eicher-Miller, Toby Jia-Jun Li, Meng Jiang, and Ronald~A Metoyer. 2025.
\newblock Limitations of the llm-as-a-judge approach for evaluating llm outputs in expert knowledge tasks.
\newblock In \emph{Proceedings of the 30th International Conference on Intelligent User Interfaces}, pages 952--966.

\bibitem[{Touvron et~al.(2023)Touvron, Martin, Stone, Albert, Almahairi, Babaei, Bashlykov, Batra, Bhargava, Bhosale et~al.}]{touvron2023llama}
Hugo Touvron, Louis Martin, Kevin Stone, Peter Albert, Amjad Almahairi, Yasmine Babaei, Nikolay Bashlykov, Soumya Batra, Prajjwal Bhargava, Shruti Bhosale, et~al. 2023.
\newblock Llama 2: Open foundation and fine-tuned chat models.
\newblock \emph{arXiv preprint arXiv:2307.09288}.

\bibitem[{Wang et~al.(2023)Wang, Wang, Mi, Wang, Xu, and Wong}]{wang2023chain}
Hongru Wang, Rui Wang, Fei Mi, Zezhong Wang, Ruifeng Xu, and Kam-Fai Wong. 2023.
\newblock Chain-of-thought prompting for responding to in-depth dialogue questions with llm.
\newblock \emph{arXiv preprint arXiv:2305.11792}.

\bibitem[{Wang et~al.(2024)Wang, Zhang, and Zhao}]{wang2024tag}
Jian Wang, Zhe Zhang, and Guosheng Zhao. 2024.
\newblock Tag-based self-learning task recommendation for mobile crowdsensing via collaborative multi-expert system.
\newblock \emph{Computer Communications}, 214:260--269.

\bibitem[{Wang et~al.(2020)Wang, Ren, Liu, Yu, and Zhang}]{wang2020qplex}
Jianhao Wang, Zhizhou Ren, Terry Liu, Yang Yu, and Chongjie Zhang. 2020.
\newblock Qplex: Duplex dueling multi-agent q-learning.
\newblock \emph{arXiv preprint arXiv:2008.01062}.

\bibitem[{Wolf et~al.(2020)Wolf, Debut, Sanh, Chaumond, Delangue, Moi, Cistac, Rault, Louf, Funtowicz, Davison, Shleifer, von Platen, Ma, Jernite, Plu, Xu, Le~Scao, Gugger, Drame, Lhoest, and Rush}]{wolf-etal-2020-transformers}
Thomas Wolf, Lysandre Debut, Victor Sanh, Julien Chaumond, Clement Delangue, Anthony Moi, Pierric Cistac, Tim Rault, Remi Louf, Morgan Funtowicz, Joe Davison, Sam Shleifer, Patrick von Platen, Clara Ma, Yacine Jernite, Julien Plu, Canwen Xu, Teven Le~Scao, Sylvain Gugger, Mariama Drame, Quentin Lhoest, and Alexander Rush. 2020.
\newblock \href {https://doi.org/10.18653/v1/2020.emnlp-demos.6} {Transformers: State-of-the-art natural language processing}.
\newblock In \emph{Proceedings of the 2020 Conference on Empirical Methods in Natural Language Processing: System Demonstrations}, pages 38--45, Online. Association for Computational Linguistics.

\bibitem[{Wu et~al.(2023)Wu, Li, Zhang, Kang, Sun, and Liu}]{wu2023community}
Yang Wu, Xurui Li, Xuhong Zhang, Yangyang Kang, Changlong Sun, and Xiaozhong Liu. 2023.
\newblock Community-based hierarchical positive-unlabeled (pu) model fusion for chronic disease prediction.
\newblock In \emph{Proceedings of the 32nd ACM International Conference on Information and Knowledge Management}, pages 2747--2756.

\bibitem[{Wu et~al.(2024{\natexlab{a}})Wu, Wang, Gumusel, and Liu}]{wu2024knowledge}
Yang Wu, Chenghao Wang, Ece Gumusel, and Xiaozhong Liu. 2024{\natexlab{a}}.
\newblock Knowledge-infused legal wisdom: Navigating llm consultation through the lens of diagnostics and positive-unlabeled reinforcement learning.
\newblock \emph{arXiv preprint arXiv:2406.03600}.

\bibitem[{Wu et~al.(2024{\natexlab{b}})Wu, Zhang, Jiao, Ma, Liu, Yu, Zhang, Yu, and Xu}]{wu2024rose}
Yang Wu, Huayi Zhang, Yizheng Jiao, Lin Ma, Xiaozhong Liu, Jinhong Yu, Dongyu Zhang, Dezhi Yu, and Wei Xu. 2024{\natexlab{b}}.
\newblock Rose: A reward-oriented data selection framework for llm task-specific instruction tuning.
\newblock \emph{arXiv preprint arXiv:2412.00631}.

\bibitem[{Xia et~al.(2024)Xia, Malladi, Gururangan, Arora, and Chen}]{xia2024less}
Mengzhou Xia, Sadhika Malladi, Suchin Gururangan, Sanjeev Arora, and Danqi Chen. 2024.
\newblock Less: Selecting influential data for targeted instruction tuning.
\newblock \emph{arXiv preprint arXiv:2402.04333}.

\bibitem[{Yao et~al.(2025{\natexlab{a}})Yao, Wu, Wang, Xiong, Wang, and Liu}]{yao2025elevating}
Rujing Yao, Yang Wu, Chenghao Wang, Jingwei Xiong, Fang Wang, and Xiaozhong Liu. 2025{\natexlab{a}}.
\newblock Elevating legal llm responses: Harnessing trainable logical structures and semantic knowledge with legal reasoning.
\newblock \emph{arXiv preprint arXiv:2502.07912}.

\bibitem[{Yao et~al.(2025{\natexlab{b}})Yao, Wu, Zhang, Zhang, Huang, Wu, Yang, Sun, Wang, and Liu}]{yao2025intelligent}
Rujing Yao, Yiquan Wu, Tong Zhang, Xuhui Zhang, Yuting Huang, Yang Wu, Jiayin Yang, Changlong Sun, Fang Wang, and Xiaozhong Liu. 2025{\natexlab{b}}.
\newblock Intelligent legal assistant: An interactive clarification system for legal question answering.
\newblock In \emph{Companion Proceedings of the ACM on Web Conference 2025}, pages 2935--2938.

\bibitem[{Zhang et~al.(2023)Zhang, Li, Ma, Zhou, and Zou}]{zhang2023llmaaa}
Ruoyu Zhang, Yanzeng Li, Yongliang Ma, Ming Zhou, and Lei Zou. 2023.
\newblock Llmaaa: Making large language models as active annotators.
\newblock \emph{arXiv preprint arXiv:2310.19596}.

\end{thebibliography}
\bibliographystyle{acl_natbib}

\nocite{}
\clearpage 

\appendix

\section{Dataset Statistics}
\label{appendix:data statictics}
In this section, we present key statistics of the CKAD dataset in Table \ref{data_overview}, which includes question–answer pairs generated from PubMed 2024 articles focused on Sepsis and Cancer NK cell mechanisms.
\begin{table}[htp]
\small
\centering
\caption{Statistics of CKAD dataset.}
\label{data_overview}
\resizebox{\linewidth}{!}{  
\setlength{\tabcolsep}{3mm}
\begin{tabular}{cc}
\hline
Disease Type & Cancer\_NK and Sepsis \\
\hline
$\# \text{Train}$ & 38,575 \\
$\# \text{Dev}$ & 4,722 \\
$\# \text{Test}$ & 4,722 \\
$\# \text{Avg.\ Tokens in Question}$ & 12 \\
$\# \text{Avg.\ Tokens in Answer}$ & 29 \\
\hline
\end{tabular}
}
\end{table}

\section{Generated Question-Answer Examples from PubMed Publications}

\begin{prompt}{Cancer QA Example from PubMed}

Question: \\
\\
What role does MHC-I play in modulating NK cell activity against tumor cells?\\
\\
Answer:\\
\\
MHC-I molecules on tumor cells can engage with inhibitory receptors on natural killer (NK) cells, such as KIRs and NKG2A, reducing NK cell activation and cytotoxic activity against the tumor.
\\
\\
\end{prompt}

\begin{prompt}{Sepsis QA Example from PubMed}

Question: \\
\\
How does anti-thrombin administration influence the risk of bleeding complications in sepsis patients?\\
\\
Answer:\\
\\
Anti-thrombin administration can increase the risk of bleeding complications by enhancing anticoagulant activity, which may impair the body's ability to form necessary clots and maintain hemostasis.
\\
\\
\end{prompt}

\section{Prompts}
\label{prompt}
In this section, we present the detailed prompts used for generating Question-Answer data and the specific prompts employed for model evaluation across all experiments. For evaluation, we slightly rephrase the final prompt instruction—asking the model to choose between output a or b—to simplify post-processing and ensure compatibility with the AlpacaEval \citep{dubois2024length} evaluation package. This modification does not affect the evaluation result, which remains to judge whether output b correctly reflects the meaning of output a. Notably, this evaluation does not simply reflect the judge model’s general preference between two answers but specifically assesses whether the target model’s answer adequately captures the core meaning of the golden answer.

\subsection{QA Extraction}

\begin{prompt}{PubMed Paper Question-Answer Generation}
\textless\textbar im\_start\textbar\textgreater system\\
You are an expert in extracting specific and relevant question-answer pairs from scientific papers. Your task is to generate five QA pairs based on the unique mechanisms or processes described in the provided paper. Focus on extracting detailed mechanisms or processes, avoiding generic or summarization-style questions.\\
\\
Guidelines: \\
1. The questions must specifically target mechanisms, processes, or detailed explanations provided in the paper. Focus on "how" or "why" certain processes or mechanisms work according to the paper. \\
2. Avoid generic or summarization-style questions, such as broad overviews or general statements about findings. \\
3. Each question should be clear, concise, and specific, addressing a mechanism, interaction, or process described in the paper. \\
4. The answers must directly explain the mechanism or process, based on specific information from the paper, and be precise and to the point. \\
\\
Examples:\\- Question 1: How does cytokine IL-15 regulate the activation of natural killer cells in the study? \\
Answer: Cytokine IL-15 regulates natural killer cell activation by binding to its receptor, triggering a signaling cascade that enhances proliferation and cytotoxic activity. \\
- Question 2: What mechanism underlies the feedback loop described for natural killer cell regulation? \\
Answer: The feedback loop involves cytokine signaling that stimulates metabolic reprogramming in natural killer cells, which in turn amplifies cytokine production. \\
\\
\textless\textbar im\_end\textbar\textgreater\\
\textless\textbar im\_start\textbar\textgreater user\\
\\
Below is the content of the paper: \\
<Insert the paper's abstract, introduction, and methodology here.> \\
Your task is to generate five QA pairs based on the unique mechanisms or processes described in the provided paper. Focus on extracting detailed mechanisms or processes, avoiding generic or summarization-style questions. The response format should be: \\
<Question: {\color{blue}The generated question>} \\
<Answer: {\color{blue}The generated answer>} \\
The generated five QA pairs are: \\
\\
\textless\textbar im\_end\textbar\textgreater\\
\end{prompt}

\subsection{Judge Prompt}
\label{prompt:gpt4_pairwise}

\begin{prompt}{Evaluation Prompt}
\textless\textbar im\_start\textbar\textgreater system\\
You are a teacher assessing whether a Output (b) correctly covers the core meaning of a Output (a) for a given Question. 
The Output (b) must fully address the question, just as the Output (a) does. Follow these rules strictly:
\\
\#\# Scoring Criteria\\
\\
1. **Semantic Match**:
   - The Output (b) must **precisely match** the meaning of the Output (a) without significant divergence.
   - Output (b) must address the Question in the same way as the Output (a).
\\
2. **Supplementary Information**:
   - Additional details are allowed **only if they do not conflict** with the Output (a).
   - Output (b) must not contain any contradictions, factual errors, or misleading information.
\\
\\
\#\# Evaluation Process\\
\\
1. **Key Point Extraction**:
   - Extract core facts, entities, and logical relationships from the Output (b).
   - Compare these with the Output (a).
   - Identify missing points, contradictory statements, or factual errors.
   - Output (b) must address the Question in the same way as the Output (a).

\textless\textbar im\_end\textbar\textgreater\\
\textless\textbar im\_start\textbar\textgreater user\\
I require an assessment of whether Output (b) correctly conveys the core meaning of Output (a). I'll provide you with a question and two model outputs. Your task is to evaluate and return either Output (a) or Output (b), based on the scoring criteria. \\

\#\# Question\\
\\
\{\\
    \verb|    |``question'': ````{\color{blue}\{Question\}}''''\\
\}\\
\\
\#\# Model Outputs\\
\\
Here are the unordered outputs from the models. Each output is associated with a specific model, identified by a unique model identifier.\\
\\
\{\\
    \verb|    |\{\\
    \verb|    |\verb|    |    ``model\_identifier'': ``m'',\\
    \verb|    |\verb|    |    ``output'': ````{\color{blue}{\{Output (a)\}}}''''\\
\verb|    |\},\\
\verb|    |\{\\
    \verb|    |\verb|    |    ``model\_identifier'': ``M'',\\
    \verb|    |\verb|    |    ``output'': ````{\color{blue}{\{Output (b)\}}}''''\\
\verb|    |\}\\
\}\\
\\
\\
\\
\#\# What's your evaluation, Output (a) or Output (b)? \\
\\
\textless\textbar im\_end\textbar\textgreater\\
\end{prompt}

\section{Expert-Wise Attention}
\label{expert_atta}

Given a question embedding $E_q^i$ and expert embeddings $E_e^j$, we define the expert-wise attention mechanism as follows:

\begin{equation}
\small
    \begin{aligned}
        e_{ij} = \sigma\left(W\cdot\left[E_q^i, E_e^j\right] + b\right)
     \label{att_expert}
    \end{aligned}
\end{equation}

\begin{equation}
\small
    \begin{aligned}
        \alpha_{ij} =
    \frac{\exp\left(\sigma\left(W\cdot\left[E_q^i, E_e^j\right] + b\right)\right)}{\sum_{k\in E_e} \exp\left(\sigma\left(W\cdot\left[E_q^i, E_e^k\right] + b\right)\right)}
    \label{att2_expert}
    \end{aligned}
\end{equation}

\begin{equation}
\small
    \begin{aligned}
        Z_i = \sum _{k\in E_e} \alpha_{ij} E_e^k
    \label{sum_expert}
    \end{aligned}
\end{equation}

where $\sigma$ denotes the $ReLU$ activation function, and $\left[.,.\right]$ represents embedding concatenation. Furthermore, we concatenate expert-wise question representation $Z_i$ with each expert embedding $E_e^j$ and pass it through an MLP $\Upsilon$ to obtain the output probability:

\begin{equation}
\small
    \begin{aligned}
        P\left(E_q^i, E_e^j\right) = \Upsilon \left(\left[Z_i, E_e^j\right]\right)
    \label{pro_expert}
    \end{aligned}
\end{equation}

\section{Overall Comparison of Question Selection Methods}
Table \ref{tab:avg_result} summarizes the average performance of different question selection methods across all expert allocation strategies. PU-ADKA shows superior results on both WR and LC\_WR metrics, with LESS ranking second in overall effectiveness. We therefore use LESS as the default question selection method for baselines in our ablation studies.
\begin{table}[htp]  
  \centering
  \caption{Average performance of different question selection methods across three expert allocation strategies.}
  \setlength{\tabcolsep}{1.8mm} 
  \resizebox{\linewidth}{!}{%
  \begin{tabular}{lccc}
    \toprule
    Method & WR.Avg (\%) & LC\_WR.Avg (\%) & Overall.Avg (\%) \\ \midrule
    RAND & 6.48 & 20.40 & 13.44 \\ 
    DEITA & \underline{10.43} & 21.40 & 15.92 \\
    CHERRY & 9.08 & 21.95 & 15.52 \\
    NUGGETS & 9.80 & 21.17 & 15.49 \\
    LESS & 10.00 & \underline{22.55} & \underline{16.27} \\
    ROSE & 9.37 & 21.95 & 15.66 \\
    \ours{} & \textbf{17.45} & \textbf{26.05} & \textbf{21.75} \\
    \bottomrule
  \end{tabular}
  }
  \label{tab:avg_result}
\end{table}

\begin{table*}[]
\caption{Performance comparison under fixed-budget and fully annotated settings. The FULL setting denotes that all available data are exhaustively annotated without any budget constraints, serving as an empirical upper bound.}
\centering
\renewcommand\arraystretch{0.97}
\label{table:full-data}
\scalebox{0.8}{
\begin{tabular}{l|l|c|c|c|c|c} 
\toprule
\multirow{2}{*}{\shortstack[l]{\textbf{Expert}\\\textbf{Allocation}}} & \multirow{2}{*}{\shortstack[l]{\textbf{Question}\\\textbf{Selection}}} & \bf GPT-4o-2024-08-06 & \bf GPT-4-Turbo & \bf GPT-4o-2024-08-06 & \bf GPT-4-Turbo & \bf Avg.Length \\ 
\cmidrule(lr){3-4} \cmidrule(lr){5-6} \cmidrule(lr){7-7}
& & WR (\%) & WR (\%) & LC\_WR (\%) & LC\_WR (\%) & - \\ \midrule

- & FULL & \textbf{22.1} \std{0.7} & \textbf{19.3} \std{0.9} & \textbf{27.8} \std{1.0} & \textbf{28.1} \std{0.8} & 1752 \\ \midrule

Random & \multirow[t]{3}{*}{LESS} 
& 7.9 \std{0.2} & 7.9 \std{0.2} & 22.0 \std{1.0} & 24.0 \std{1.1} & 2212 \\ 
Cost-Greedy &  & 12.1 \std{0.4} & 9.6 \std{0.4} & 22.1 \std{0.8} & 21.2 \std{1.0} & 2218 \\
Match-Greedy &  & 12.1 \std{0.4} & 10.4 \std{0.2} & 23.5 \std{1.0} & 22.5 \std{1.0} & 2252 \\ \midrule

Ours & \ours{} & \underline{18.2} \std{0.6} & \underline{16.7} \std{0.4} & \underline{25.6} \std{1.0} & \underline{26.5} \std{0.9} & 1781 \\ 
\bottomrule
\end{tabular}
}
\end{table*}

\begin{table*}[]
\caption{Performance comparison between different encoders used for question representation within PU-ADKA framework.}
\centering
\renewcommand\arraystretch{0.97}
\label{tb:encoder}
\scalebox{0.8}{
\begin{tabular}{l|c|c|c|c|c} 
\toprule
\multirow{2}{*}{\shortstack[l]{\textbf{Encoder}}} & \bf GPT-4o-2024-08-06 & \bf GPT-4-Turbo & \bf GPT-4o-2024-08-06 & \bf GPT-4-Turbo & \bf Avg.Length \\ 
\cmidrule(lr){2-3} \cmidrule(lr){4-5} \cmidrule(lr){6-6}
& WR (\%) & WR (\%) & LC\_WR (\%) & LC\_WR (\%) & - \\ \midrule

BERT-base & \underline{16.3} \std{0.9} & \underline{12.9} \std{0.7} & \underline{24.0} \std{1.0} & \underline{25.4} \std{1.2} & 1967 \\ \midrule

Llama2-7B (ours) & \textbf{18.2} \std{0.6} & \textbf{16.7} \std{0.4} & \textbf{25.6} \std{1.0} & \textbf{26.5} \std{0.9} & 1781 \\ 
\bottomrule
\end{tabular}
}
\end{table*}

\section{Comparison with Fully Annotated Upper Bound}
To understand the upper performance bound achievable without cost constraints, we compare PU-ADKA against a setting where all training data are fully annotated (FULL). As expected, FULL yields the highest scores in Table~\ref{table:full-data}, serving as an empirical upper bound. \begin{table}[htp]  
  \centering
  \caption{Number of annotated QA pairs and evaluation results under different expert allocation strategies (questions selected by LESS; judged by GPT-4o-2024-08-06).}
  \setlength{\tabcolsep}{1.8mm}
  \resizebox{\linewidth}{!}{%
  \begin{tabular}{lccc}
    \toprule
    Method & Annotated QA Pairs & WR (\%) & LC\_WR (\%) \\ \midrule
    Random & 312 & 7.9 & 22.0 \\
    Cost-Greedy & 1000 & \underline{12.1} & 22.1 \\
    Match-Greedy & 508 & \underline{12.1} & \underline{23.5} \\
    Ours (PU-ADKA) & 632 & \textbf{18.2} & \textbf{25.6} \\
    \bottomrule
  \end{tabular}
  }
  \label{tab:expert_allocation}
\end{table}
However, PU-ADKA approaches this upper limit while operating under a strict \$100 budget—substantially outperforming all baselines in both WR and LC\_WR scores. This highlights the effectiveness of PU-ADKA in achieving strong domain adaptation with fewer annotations.

\section{Effect of Encoder Architecture on Performance}
To assess the impact of the encoder architecture on our framework, we compare a standard BERT-base \citep{devlin2019bert} model with our default Llama2-7B encoder. Results in Table~\ref{tb:encoder} show that Llama2-7B consistently outperforms BERT-base encoder across all evaluation metrics.

\section{Annotation Quantity under Budget Constraints}
Table~\ref{tab:expert_allocation} shows that although the Cost-Greedy strategy yields the largest number of annotated QA pairs, it does not achieve the best performance. \begin{table}[h]
\centering
\small
\caption{Data Quality Scoring Form}
\label{tab:quality_rubric}
\begin{tabular}{c|p{6cm}}
\hline
Score & Description \\
\hline
1 & Incorrect or irrelevant. \\
2 & Partially correct, key issues. \\
3 & Correct and main point covered. \\
4 & Correct with minor omissions. \\
5 & Fully correct and complete. \\
\hline
\end{tabular}
\end{table}In contrast, PU-ADKA achieves a balance between annotation quantity and quality: it produces more annotations than Random and Match-Greedy strategies and achieves the highest WR and LC\_WR scores. This demonstrates PU-ADKA’s effectiveness in utilizing limited budgets to acquire high-quality supervision.

\section{Data Quality Scoring Form}
\label{data_quality_rubric}

The quality of each QA pair is scored on a 1–5 scale based on the correctness and completeness of its biomedical mechanistic explanation. The scoring rubric is shown in Table~\ref{tab:quality_rubric}.

\section{Discussion}

We adopt the base Llama2-7B model rather than an instruction-tuned variant to ensure a controlled setting where observed improvements are attributable solely to our expert-interaction framework, without influence from pretrained instruction-following capabilities. Besides, GPT-4o-2024-08-06 is used to generate QA pairs based on the factual content of 2024 PubMed articles. This generation process is independent of any model comparison or evaluation. To mitigate potential bias from using a single evaluation model, we also include GPT-4-Turbo as a second judge model.

\end{document}